\pdfoutput=1

\documentclass[11pt]{article}
\usepackage{authblk}
\usepackage[]{acl}

\usepackage{times}
\usepackage{latexsym}

\usepackage{tipa}

\usepackage[T2A,T1]{fontenc}
\usepackage[utf8]{inputenc}
\usepackage[russian,english]{babel}
\usepackage{booktabs}

\usepackage{microtype}
\usepackage{amsmath}
\usepackage{graphicx}
\usepackage{subfigure}
\usepackage{amsfonts}
\usepackage{multirow}
\usepackage{listings}
\usepackage{url}
\usepackage[ruled,linesnumbered]{algorithm2e}  
\usepackage{CJKutf8}
\usepackage{graphicx}
\usepackage{subfigure}
\usepackage{supertabular}
\usepackage{subfigure}
\usepackage{dcolumn,booktabs}
\usepackage{amssymb}

\title{On the Copying Problem of Unsupervised NMT: \\
A Training Schedule with a Language Discriminator Loss}

\author[*$\diamond$]{\bf Yihong Liu}
\author[*$\diamond$]{\bf Alexandra Chronopoulou}
\author[*$\diamond$]{\bf Hinrich Sch\"utze}
\author[*$\diamond$]{\bf Alexander Fraser}

\affil[*]{Center for Information and Language Processing, LMU Munich} \affil[$\diamond$]{Munich Center for Machine Learning (MCML)
 \protect\\ \texttt{\{yihong, achron, fraser\}@cis.lmu.de}}

\newcounter{notecounter}
\newcommand{\enotesoff}{\long\gdef\enote##1##2{}}
\newcommand{\enoteson}{\long\gdef\enote##1##2{{
\stepcounter{notecounter}
{\large\bf \hspace{1cm}\arabic{notecounter} $<<<$ ##1: ##2 $>>>$\hspace{1cm}}}}}
\enoteson
\enotesoff

\newcommand{\veryshortarrow}[1][3pt]{\mathrel{%
   \hbox{\rule[\dimexpr\fontdimen22\textfont2-.2pt\relax]{#1}{.4pt}}%
   \mkern-4mu\hbox{\usefont{U}{lasy}{m}{n}\symbol{41}}}}

\begin{document}
\maketitle

\begin{abstract}
Although unsupervised neural machine translation (UNMT) has achieved success in many language pairs, the copying problem, i.e., directly copying some parts of the input sentence as the translation, is common among distant language pairs, especially when low-resource languages are involved. We find this issue is closely related to an unexpected copying behavior during online back-translation (BT). In this work, we propose a simple but effective training schedule that incorporates a language discriminator loss. The loss imposes constraints on the intermediate translation so that the translation is in the desired language. By conducting extensive experiments on different language pairs, including similar and distant, high and low-resource languages, we find that our method alleviates the copying problem, thus improving the translation performance on low-resource languages. 
\end{abstract}

\section{Introduction}

UNMT \citep{LampleCDR18umtmco, ArtetxeLAC18unmt} is a new and effective approach for tackling the scarcity of parallel data. Typically, a cross-lingual pretrained language model (PLM) \citep{petersetal2018deep, devlinetal2019bert} is trained on two languages and then used to initialize the model for 
%
the
UNMT task \citep{Conneau2019xlm, SongTQLL19mass, yangetal2020csp, liuetal2020multilingualdenoising}. However, when it comes to low-resource languages, especially when translating between distant language pairs, UNMT often yields very poor results \citep{neubighu2018rapid, guzmanetal2019flores, marchisioetal2020unsupervised}. One of the major problems that lead to low translation quality is
the
copying problem or off-target problem \citep{kimetal2020unsupervised,zhangetal2020improving}. That is: the trained model does not translate but copies some words or even the whole sentence from the input as the translation.

We find the copying problem is closely related to an unexpected behavior in BT \citep{sennrichetal2016improving}: the model does not translate into the correct intermediate language but simply copies tokens from the source language. To address this problem, this work proposes a simple but effective method that can be integrated into the standard UNMT training. We leverage a language discriminator to detect the language of the intermediate translation generated in BT and backpropagate the gradients to the main model. In this way, we can provide implicit supervision to the model. We find that by adding such a training objective, the copying problem can be largely alleviated, especially for low-resource languages. Noticeably, we do not introduce any language-specific architectures into the main model. To the best of our knowledge, this is the first work that introduces a language discriminator loss to force the intermediate translations in BT to be in the correct language. The contributions of our work are as follows:\\
(1) We explore the reasons behind the copying problem in UNMT and propose a training schedule with a language discriminator loss.\\
(2) We evaluate 
our method on many languages, including high- and low-resource, and similar and distant language pairs.\\
(3) We carry out an analysis, showing the proposed method can reduce the copying ratio, especially on small-size datasets and distant language pairs. \\
(4) We make our code publicly available. \footnote{{\url{https://github.com/yihongL1U/xlm_lang_dis}}}

\section{Problem Statement \& Approach}

\subsection{Copying Problem}\label{sec:copying}
The copying problem is also known as an off-target translation issue in multilingual NMT especially zero-shot scenario \citep{guetal2019improved,yang2021improving,chen2023off}. One important task in zero-shot NMT is to let the model translate into the correct language given so many target languages. Our motivation in UNMT is similar, while each UNMT model often specifically deals with two languages, therefore only two translation directions are considered. Although adding language tags \citep{wu-etal-2021-language} is effective in addressing the copying problem in multilingual NMT, it is not a standard process in UNMT. This is because a language embedding is often added to each token embedding \citep{Conneau2019xlm,SongTQLL19mass,liuetal2022flow}. Language embeddings have similar functions to language tags: providing information about the language of each token. Unfortunately, language embeddings turn out to be not very effective in addressing the copying problem, especially for low-resource or distant language pairs \citep{kimetal2020unsupervised}. Thus, in this work, we explore why the copying problem occurs and how we can alleviate it in UNMT. We analyze the problem from two perspectives:\\

\begin{figure}
  \centering
  \includegraphics[width=0.48\textwidth]{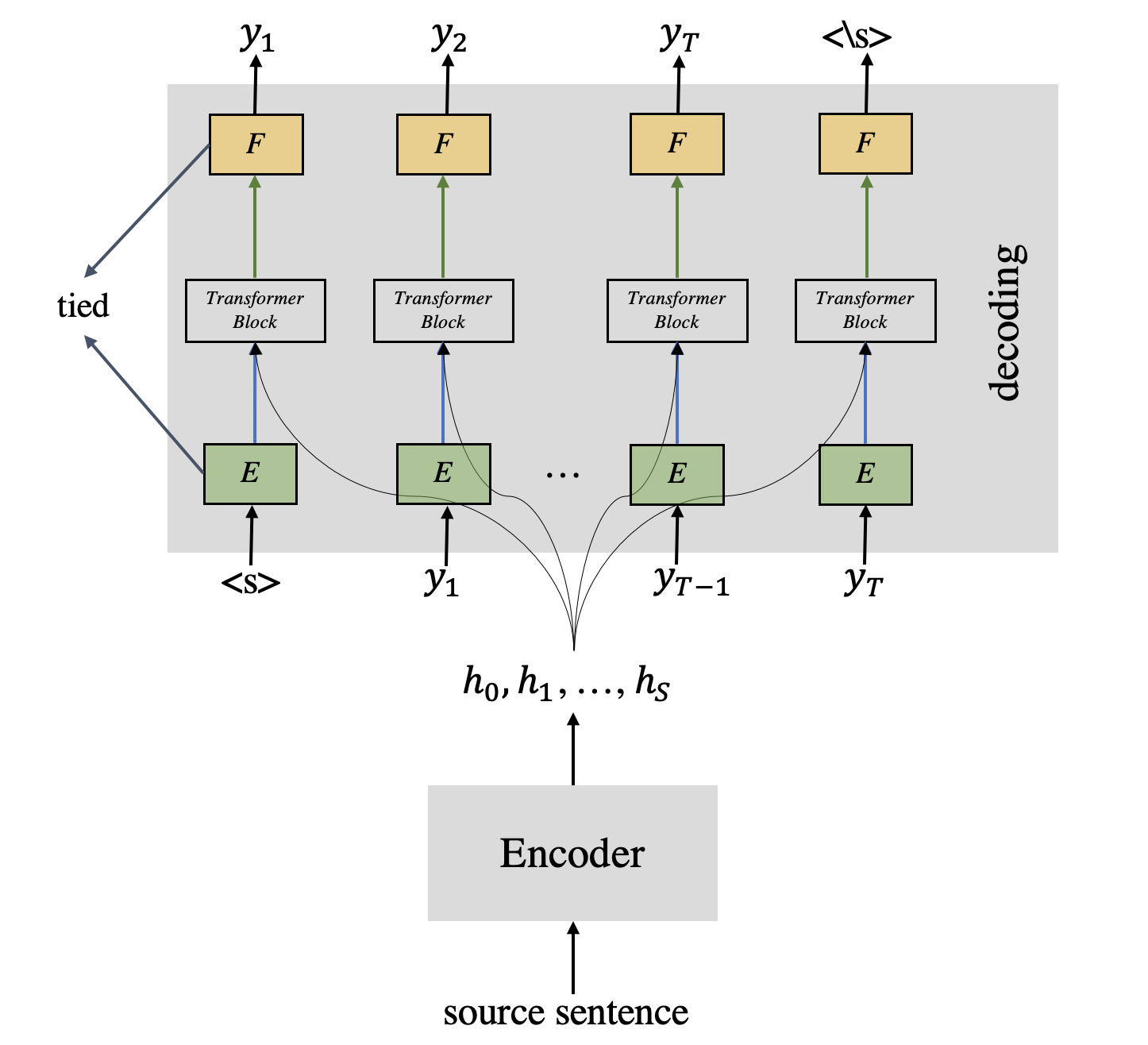}
  \caption{A view of the UNMT architecture. The weights of the final fully connected layer (block \textit{F}) are tied with the weight of the embedding layer ( block \textit{E}).} \label{fig:decoding}
\end{figure}

\begin{figure}
  \centering
  \includegraphics[width=0.48\textwidth]{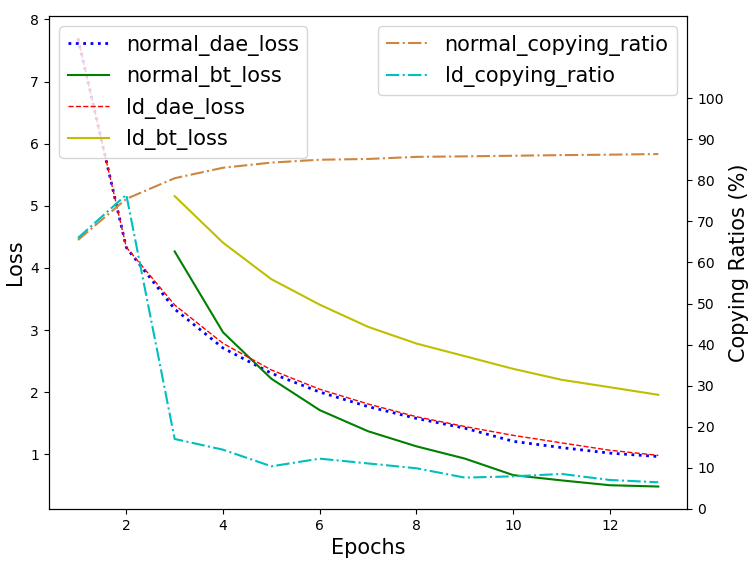}
  \caption{The losses (left ordinate) and copying ratios (right ordinate) of Multi30K English-French pair over epochs. The normal\_dae\_loss (resp. normal\_bt\_loss) and normal\_copying\_ratio are DAE loss (resp. BT loss) and copying ratio from the \textbf{vanilla} UNMT. The ld\_dae\_loss (resp. ld\_bt\_loss) and ld\_copying\_ratio are DAE loss (resp. BT loss) and the copying ratio from the UNMT incorporated with the \textbf{language discriminator}.}\label{fig:losses}
\end{figure}

\noindent \textbf{Architecture perspective.} In UNMT, the weight of the final fully connected layer (for obtaining the logits of each word in the vocabulary) is often tied to the weight of a cross-lingual embedding layer, as shown in Figure \ref{fig:decoding}. That is, the representations of tokens from two languages are shared in the same space. Although this setting is arguably a better starting point for most modern NMT models, it unfortunately also allows the models to generate a token in an unexpected language at any time step. Furthermore, because of an autoregressive decoder, errors can easily accumulate, as the tokens initially generated by the model highly influence the generation of the subsequent tokens. In contrast to this setting, using separate word look-up tables or separate decoders for involved languages can address the problem \citep{LampleCDR18umtmco,liuetal2022flow}. However, such a setting can be harmful for learning cross-lingual knowledge and largely increase the number of parameters. In this view, it is desired to keep the structure simple (no language-specific architecture) while preventing the model from decoding in a copying way.\\
\\
\noindent \textbf{Objective perspective.} Typically, a UNMT model is trained by denoising autoencoding (DAE) \citep{vincent2008extracting} and online back-translation (BT) \citep{sennrichetal2016improving} objectives. In DAE objective, even though the model is trained to denoise on two languages simultaneously, there is no guarantee that the model can transfer the cross-lingual information that might improve translation between the two languages. In fact, \citet{SongTQLL19mass} empirically find that a pretrained encoder-decoder model with DAE objective can even perform worse than the model without it because DAE encourages the model to perform the copying. In comparison with DAE, BT is arguably more important, as it tries to directly optimize the translation. However, we find that BT can also “fail” during training. That is, the model can take the shortcut, i.e., copy the input sentence as the intermediate translation and then copy it again for the reconstruction. By taking such a shortcut, the loss of BT can quickly decrease while the copying ratio \citep{liuetal2021copying}, a metric to measure the percentage of generated tokens that are copied from the input, keeps increasing and reaches a high-value plateau, as shown in Figure \ref{fig:losses}.
This indicates that: because of no constraints on the intermediate translation, the model can always choose the easiest shortcut for BT, which finally corrupts the model's translation capability.

\subsection{A Language Discriminator Loss}\label{sec:LD_details}

To avoid such an unexpected copying behavior in BT, our intuition suggests that forcing the intermediate generation to be in the correct language would be helpful. Instead of forcing all tokens, we could simply force the first token to be in the correct language, because the first generated token will influence the generation of all the subsequent tokens.
Next, the problem is how to force the first generated token to be in the desired target language. An equivalent question would be: \emph{how can we force the output vector of the decoder at the first time step to be closer to the embedding of a token in the target language?} The answer might be trivial. We could use a trained \textbf{language discriminator} (LD), which is a classifier, to classify the first-time-step output vectors of the decoder and then backpropagate the gradients to the main model (encoder and decoder). In this way, the model knows which intermediate language it should generate for the first-time-step token, therefore preventing the copying behavior.

For training LD, we could use the first-time-step outputs of the decoder in DAE steps. The LD is trained to predict the language of the first-time-step outputs by minimizing the cross entropy loss:
\begin{equation}
\mathcal{L}_{LD} = \mathbb{E}_{x \sim \mathcal{D}_l}[p(l | LD(\mathcal{O}_l)]
\end{equation}
where $LD$ is the language discriminator, $\mathcal{O}_l$ are the first-time-step outputs generated  by $Dec(Enc(x, l), l)$ and $l$ denotes the language (either \textit{src} or \textit{tgt}).
Notably, $\mathcal{L}_{LD}$  only backpropagates to the language discriminator in the DAE step. In this way, the discriminator is able to distinguish representations from different languages.

In the BT process, the language discriminator is fixed and $\mathcal{L}_{LD}$ loss is only used to update the main model so it learns to differentiate representations from different languages. Taking \textit{src-tgt-src} BT for example, the loss is as follows:
\begin{equation}
\mathcal{L}_{LD} = \mathbb{E}_{x \sim \mathcal{D}_{src}}[p(tgt | LD(\mathcal{O}_{tgt})]
\end{equation}
where $\mathcal{O}_{tgt}$ are the first-time-step outputs generated in the \textit{src}-to-\textit{tgt} step, i.e., $Dec(Enc(x, src), tgt)$. The language discriminator does not have to be used for the next step in BT, i.e., \textit{tgt}-to-\textit{src} translation, because there are already ground-truth $src$-language sentences as supervision. All we need 
to do
is to make sure the intermediate translation is in the correct language. We use a weight $\lambda_{LD}$ to control the contribution of the LD loss to the final loss that is used to update the parameters of the main model. It is easy to note that the larger the weight, the model will be more focusing on the task of distinguishing representations from different languages.

This training schedule 
is
similar to the adversarial loss \citep{Goodfellow2014GAN} used by \citet{LampleCDR18umtmco}, where they trained a discriminator to make the \textbf{outputs of the encoder} language-agnostic, aiming to improve the cross-linguality of a shared encoder. Our aim, however, is different: we want to enable the \textbf{decoder} to generate distinguishable outputs which correctly correspond to the language 
that the model is expected to generate in the BT process. Algorithm \ref{alg:dae-bt-ld} presents the training schedule in detail.

\begin{algorithm}
    \caption{Training Schedule}  
    \footnotesize
    \label{alg:dae-bt-ld}  
    \KwIn{pretrained encoder $Enc$ and decoder $Dec$, language discriminator $LD$, source and target monolingual data $\mathcal{D}_{src}$, $\mathcal{D}_{tgt}$, maximum finetuning steps $T$ and coefficient $\lambda_{LD}$ \;}  
    \KwOut{Finetuned encoder $Enc$ and decoder $Dec$)\;}
    $t \leftarrow 0$\;  
    \While{not converged \rm{or} $t < T$}   
    {
    // \textbf{for \textit{src} language do DAE and BT:}\\
    $\mathcal{B}_{src} \leftarrow$ sample batch from $\mathcal{D}_{src}$\;
    // \textit{DAE step (below)} \\
    $\mathcal{\tilde B}_{src},  \mathcal{O}_{src} \leftarrow $ generate reconstructions and first-time-step outputs from $Dec(Enc(noise(\mathcal{B}_{src}), src), src)$\;
    detach $\mathcal{O}_{src}$ from the compute graph \;
    $\boldsymbol{\theta}_{Enc},  \boldsymbol{\theta}_{Dec} \leftarrow \arg \min \mathcal{L}_{DAE} (\mathcal{B}_{src}, \mathcal{\tilde B}_{src})$\;
     $\boldsymbol{\theta}_{LD}  \leftarrow \arg \min \mathcal{L}_{LD} (\mathcal{O}_{src}, src) $\;
    // \textit{BT step (below)} \\
    freeze $\boldsymbol{\theta}_{LD}$\;
    $\mathcal{\tilde B}_{tgt},  \mathcal{O}_{tgt} \leftarrow $ generate \textit{tgt}-language translations and first-time-step outputs from $Dec(Enc(\mathcal{B}_{src}, src), tgt)$ \;
    $\mathcal{\tilde B}_{src} \leftarrow $ generate \textit{src}-language back-translations from $Dec(Enc(\mathcal{\tilde B}_{tgt}, tgt), src)$ \;
    $\boldsymbol{\theta}_{Enc},  \boldsymbol{\theta}_{Dec} \leftarrow \arg \min \mathcal{L}_{BT} (\mathcal{B}_{src}, \mathcal{\tilde B}_{src}) + \lambda_{LD} \, \mathcal{L}_{LD}(\mathcal{O}_{tgt}, tgt)$\;
    // \textbf{for \textit{tgt} language do the 
    same
    as above} \\
    $t \leftarrow t + 1$\; 
    }
    return $Enc$ and  $Dec$\;  
\end{algorithm} 

\begin{figure*}[htbp]
\centering
\subfigure[$\lambda_{LD}=0$]{
\begin{minipage}[t]{0.3\linewidth}
\centering
\includegraphics[width=1.6in]{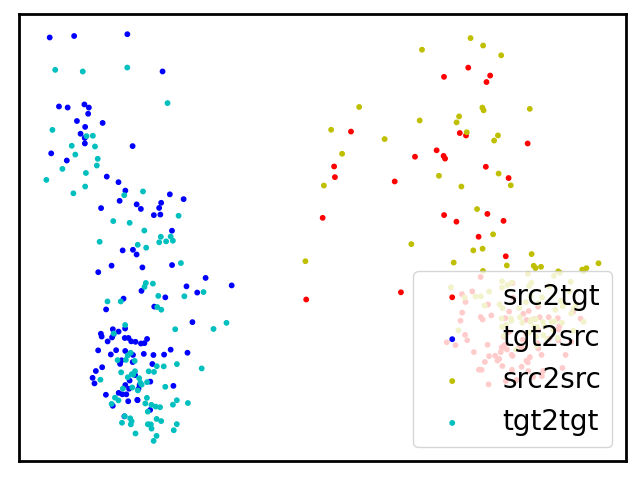}
\end{minipage}%
}%
\subfigure[$\lambda_{LD}=0.01$]{
\begin{minipage}[t]{0.3\linewidth}
\centering
\includegraphics[width=1.6in]{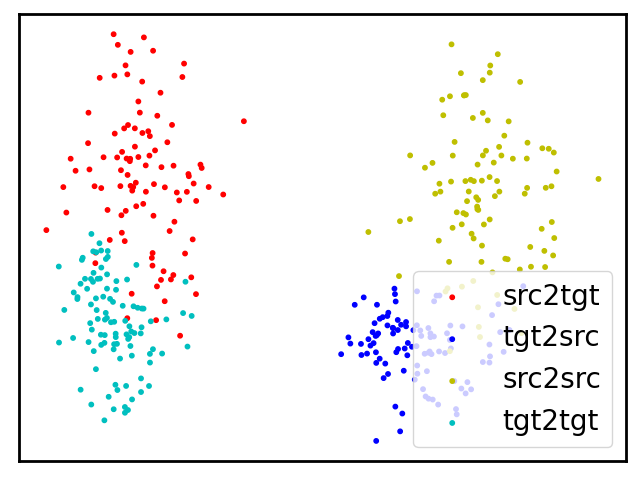}
\end{minipage}%
}%
\subfigure[$\lambda_{LD}=0.1$]{
\begin{minipage}[t]{0.3\linewidth}
\centering
\includegraphics[width=1.6in]{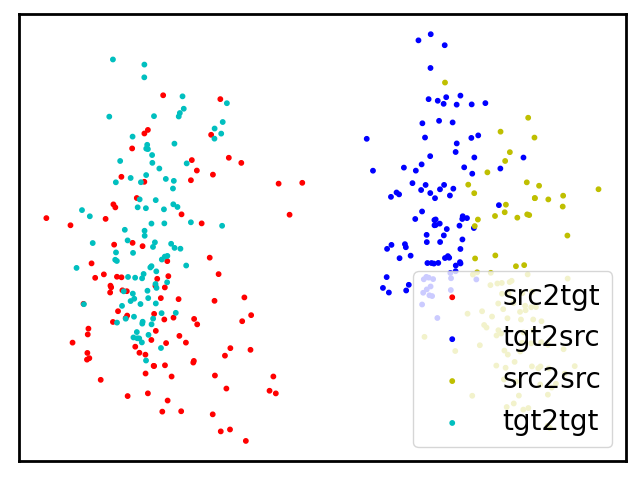}
\end{minipage}%
}%

\subfigure[$\lambda_{LD}=1$]{
\begin{minipage}[t]{0.3\linewidth}
\centering
\includegraphics[width=1.6in]{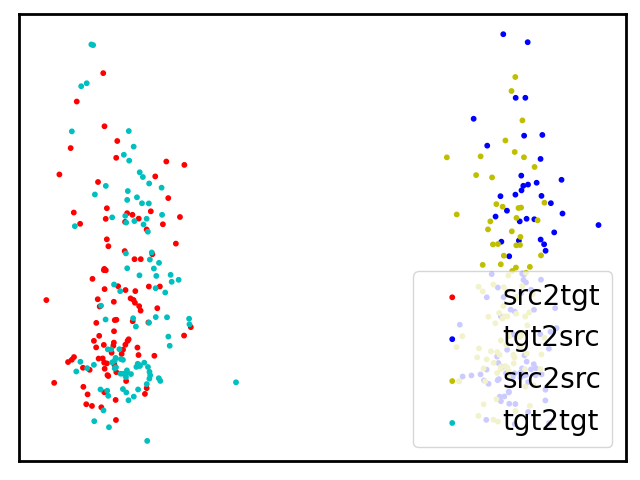}
\end{minipage}%
}%
\subfigure[$\lambda_{LD}=10$]{
\begin{minipage}[t]{0.3\linewidth}
\centering
\includegraphics[width=1.6in]{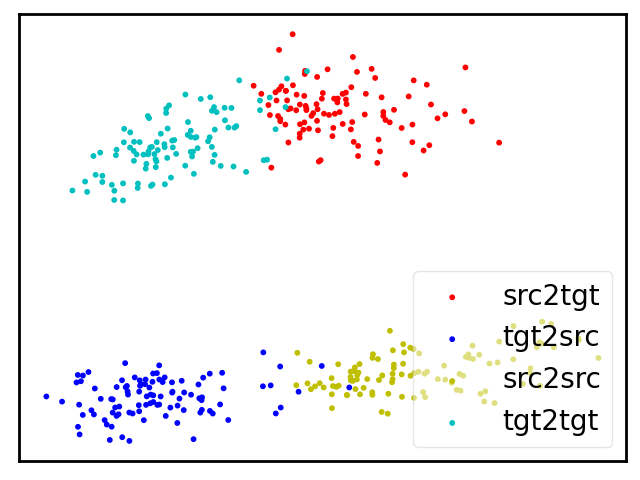}
\end{minipage}%
}%
\subfigure[$\lambda_{LD}=100$]{
\begin{minipage}[t]{0.3\linewidth}
\centering
\includegraphics[width=1.6in]{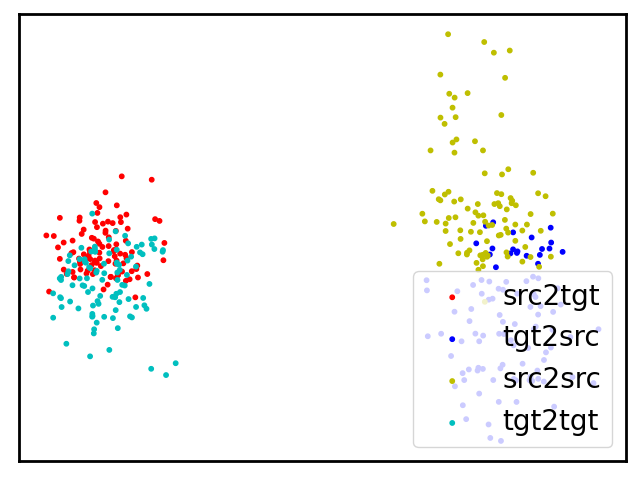}
\end{minipage}%
}%
\centering
\caption{The visualizations of the first-time-step output vectors of the decoder in UNMT trained with different weights for the proposed language discriminator loss. The dimension of the outputs is originally 1024. Principal component analysis (PCA) is leveraged to project those outputs into a 2-dimensional subspace for convenience of visualization. src2src (resp. tgt2tgt) denotes the output in the English-to-English (resp. German-to-German) autoencoding task. src2tgt (resp. tgt2src) denotes the output in the English-to-German (resp. German-to-English) translation task. The sentences used for the visualizations are the same or the corresponding parallel translations.}
\label{fig:outputs}
\end{figure*}

\section{Experiments}

\subsection{Setups}
\paragraph{Multi30K} \citep{elliott2016multi30k,elliott2017findings}\footnote{\url{https://github.com/multi30k/dataset}}. The officially provided
train, validation and test sets in English (\textbf{En}), German (\textbf{De}) and French (\textbf{Fr}) are used. Similar to \citet{LampleCDR18umtmco}, 
we only
use the caption of each image, and we split the train and validation sets into monolingual corpora by only using one-half of the data for a language.

\paragraph{WMT} \citep{barraultetal2019findings}.
We select 50M sentences for high-resource languages: English (\textbf{En}), French (\textbf{Fr}), German (\textbf{De}), Russian (\textbf{Ru}) and  Chinese (\textbf{Zh}) (14M available) and all available monolingual sentences for low-resource language: Gujarati (\textbf{Gu}) (3M), Kazakh (\textbf{Kk}) (4M). We report the results on \textit{newtest2014} for En-Fr pair,  \textit{newtest2016} for En-De pair, \textit{newtest2018} for En-Ru pair and \textit{newtest2019} for 
the remaining language
pairs. 

\paragraph{Pretrained Models}
We use cross-lingual pretrained language model (\textit{xlm-mlm-ende-1024} and \textit{xlm-mlm-enfr-1024}) from
HuggingFace\footnote{\url{https://github.com/huggingface}} \citep{wolf2020transformers} to initialize a shared encoder (parameters are fixed) in Multi30K experiments. In those experiments, we randomly initialize a shared decoder because Multi30k is so small that a randomly initialized decoder can work already very well based on our preliminary experiments. For WMT experiments, we pretrain our own cross-lingual language models using the code base of XLM\footnote{\url{https://github.com/facebookresearch/XLM}} and use the pretrained models to initialize both the encoder and decoder for UNMT task.\footnote{Details of hyperparameters and relevant information of all the models are shown in Section \ref{hyperparam} in the Appendix.}

\subsection{Analysis on Multi30K}\label{sec:multi30k}

To figure out how the LD loss could influence the performance, we use six different weights for it: 0, 0.01, 0.1, 1, 10 and 100. When the weight equals 0, the UNMT training will not consider the LD loss at all and this setting would then be exactly the same as the vanilla (i.e., DAE + BT) UNMT. The results are shown in Table \ref{tab:multi30k_perf}. 
In addition to BLEU scores \citep{papineni2002bleu}, we also compute copying ratios \citep{liuetal2021copying} for each listed direction.

\begin{table*}
\resizebox{\textwidth}{!}{ 
\begin{tabular}{c|c|c|c}
\hline
Model & Source input                                                                                                                               & Model output                                                                                                                        & \multicolumn{1}{l}{Reference output}                                                                                                   \\ \hline
$\lambda_{LD}=0$     & \multirow{6}{*}{\begin{tabular}[c]{@{}c@{}}a man in an orange hat\\  starring at something.\end{tabular}}                                  & \begin{tabular}[c]{@{}c@{}}a man in an orange hat \\ staring at something.\end{tabular}                                             & \multirow{6}{*}{\begin{tabular}[c]{@{}c@{}}ein mann mit einem \\ orangefarbenen hut, \\ der etwas anstarrt.\end{tabular}}              \\ \cline{1-1} \cline{3-3}
$\lambda_{LD}=0.01$  &                                                                                                                                            & \begin{tabular}[c]{@{}c@{}}ein mann in an orange hat \\ starring at something.\end{tabular}                                         &                                                                                                                                        \\ \cline{1-1} \cline{3-3}
$\lambda_{LD}=0,1$   &                                                                                                                                            & \begin{tabular}[c]{@{}c@{}}ein mann in an orange hat\\  gerade etwas bei etwas.\end{tabular}                                        &                                                                                                                                        \\ \cline{1-1} \cline{3-3}
$\lambda_{LD}=1$    &                                                                                                                                            & \begin{tabular}[c]{@{}c@{}}ein mann in einem orangefarbenen\\  hut spielt bei etwas.\end{tabular}                                   &                                                                                                                                        \\ \cline{1-1} \cline{3-3}
$\lambda_{LD}=10$   &                                                                                                                                            & \begin{tabular}[c]{@{}c@{}}ein mann in einem orangefarbenen\\  hut spielt bei etwas.\end{tabular}                                   &                                                                                                                                        \\ \cline{1-1} \cline{3-3}
$\lambda_{LD}=100$  &                                                                                                                                            & \begin{tabular}[c]{@{}c@{}}eine frau in einem orangefarbenen \\ hut spielt bei etwas.\end{tabular}                                  &                                                                                                                                        \\ \hline
$\lambda_{LD}=0$    & \multirow{6}{*}{\begin{tabular}[c]{@{}c@{}}a boston terrier is running \\ on lush green grass \\ in front of a white fence.\end{tabular}} & \begin{tabular}[c]{@{}c@{}}a boston dog is running on leafy grass\\  in front of a white fence. \end{tabular}                        & \multirow{6}{*}{\begin{tabular}[c]{@{}c@{}}ein boston terrier läuft \\ über saftig-grünes gras\\  vor einem weißen zaun.\end{tabular}} \\ \cline{1-1} \cline{3-3}
$\lambda_{LD}=0.01$  &                                                                                                                                            & \begin{tabular}[c]{@{}c@{}}ein boston terrier läuft auf einem gepflasterten \\ grünen grass in front of a white fence.\end{tabular} &                                                                                                                                        \\ \cline{1-1} \cline{3-3}
$\lambda_{LD}=0.1$  &                                                                                                                                            & \begin{tabular}[c]{@{}c@{}}ein boston terrier läuft auf einem grünen rasen \\ vor einem weißen zaun.\end{tabular}                   &                                                                                                                                        \\ \cline{1-1} \cline{3-3}
$\lambda_{LD}=1$     &                                                                                                                                            & \begin{tabular}[c]{@{}c@{}}ein boston terrier läuft auf einem grünen rasen\\  vor einem weißen zaun.\end{tabular}                   &                                                                                                                                        \\ \cline{1-1} \cline{3-3}
$\lambda_{LD}=10$    &                                                                                                                                            & \begin{tabular}[c]{@{}c@{}}ein boston terrier läuft auf einem grünen gras\\  vor einem weißen zaun.\end{tabular}                    &                                                                                                                                        \\ \cline{1-1} \cline{3-3}
$\lambda_{LD}=100$  &                                                                                                                                            & \begin{tabular}[c]{@{}c@{}}eine boston terrier läuft auf grünen gras\\  vor einem weißen zaun.\end{tabular}                         &                                                                                                                                        \\ \hline
\end{tabular}
}
\caption{
Examples of translations from the model trained on Multi30K dataset (En-De pair) with different weights $\lambda_{LD}$ for language discriminator loss. We do not use beam search to generate these translations.}
\label{tab:translations}
\end{table*}

\begin{table}
\centering
\resizebox{\columnwidth}{!}
{
\begin{tabular}{rrrrr}
\hline
Models & En $\veryshortarrow$ De & De $\veryshortarrow$ En & En $\veryshortarrow$ Fr & Fr $\veryshortarrow$ En \\
\hline
$0$ & 0.22 (87\%) & 0.19 (84\%) & 0.14 (89\%) & 0.10 (83\%)\\
$0.01$ & 15.78 (42\%) & 22.04 (24\%) & 24.73 (24\%) & 22.15 (25\%)\\
$0.1$ & 25.91 (14\%) & 28.46 (15\%) & 39.72 (6\%) & 37.50 (7\%)\\
$1$& \textbf{27.96} (12\%) & \textbf{30.05} (12\%) & \textbf{42.74} (5\%) & \textbf{39.02} (6\%)\\
$10$ & 24.35 (14\%) & 25.60 (13\%) & 41.26 (5\%) & 37.61 (6\%) \\
$100$ & 20.66 (12\%) & 26.74 (10\%) & 30.65 (5\%) & 32.10 (7\%) \\
\hline
\end{tabular}
}
\caption{\label{tab:multi30k_perf}
BLEU scores and copying ratios (inside parentheses) of models trained with different weights $\lambda_{LD}$ on Multi30K dataset. When the weight $\lambda_{LD}=0$, the model 
degenerates to the vanilla UNMT model.
}
\end{table}

The general trend shows that: when $0 \le \lambda_{LD} \le 1$, the BLEU scores increase and the copying ratios decrease 
when
increasing the weight, suggesting the copying problem is alleviated by introducing the LD loss. However, when $\lambda_{LD} 
>
1$, the BLEU scores decrease while copying ratios remain at the same level with the increase of the weight. This indicates that the model is over-emphasizing distinguishing the outputs when the weights are large. Therefore, moderate weights, e.g., 1, might be optimal if we want to alleviate the copying problem while achieving good translation performance.

When $\lambda_{LD} = 0$, poor BLEU scores are obtained because of the copying problem. We see that all copying ratios in Table \ref{tab:multi30k_perf} are very high: more than 80\% for all directions. Example translations from the translation model for En-De pair in Table \ref{tab:translations} show that when $\lambda_{LD} = 0$, the MT system simply copies the input sentences. It is very clear that with the increase of the weight, it becomes less likely for the model to copy the words from the source input as the output translation. However, when the weight is too large, e.g., $\lambda_{LD}=100$, there are obvious mistakes made by the translation model. For example, ``man'' in English is wrongly translated to ``frau'' (means woman) in German, ``a'' is wrongly translated into ``eine'' since boston terrier is a masculine instead of a feminine noun. Moderate weights, e.g., $\lambda_{LD} = 1$, achieves the best performance while obtaining fewer errors. 

To figure out how the LD loss influences the representations, i.e., the first-time-step output vectors generated by the decoder, we visualize these vectors in 2D by using principal component analysis (PCA), as shown in Figure \ref{fig:outputs}. The visualization verifies the relationship between the output and the occurrence of the copying problem.
src2tgt and tgt2tgt first-time-step outputs should be close to each other in the subspace as they are both used to directly generate target-language sentences. However, in  Fig. \ref{fig:outputs} (a), when $\lambda_{LD} = 0$, src2tgt and src2src are located together while tgt2src and tgt2tgt are together. In contrast, when LD loss is imposed, e.g., $\lambda_{LD} = 1$ (Fig. \ref{fig:outputs} (d)), the outputs are distributed as we expect: src2tgt and tgt2tgt are located together and tgt2src and src2src together.

\begin{table*}
\centering
\begin{tabular}{lrrrrrrrr}
\hline
Models & En $\veryshortarrow$ De & De $\veryshortarrow$ En & En $\veryshortarrow$ Fr & Fr $\veryshortarrow$ En & En $\veryshortarrow$ Ru & Ru $\veryshortarrow$ En & En $\veryshortarrow$ Zh & Zh $\veryshortarrow$ En\\
\hline
XLM baseline & \textbf{20.51} & \textbf{25.99} & \textbf{22.87} & 25.88 & \textbf{14.10} & \textbf{16.92} & 6.36 & 4.28\\
XLM (+ LD) & 20.40 & 25.85 & 21.22 & \textbf{26.92} & 13.49 & 16.12 & \textbf{6.80} & \textbf{4.69} \\
\hline
\end{tabular}
\caption{\label{tab:high_wmt_perf}
BLEU scores of the XLM baseline and the same model enhanced with the LD loss on high-resource language pairs. The scores of baseline are obtained by reproducing the published code \citep{Conneau2019xlm}. 
}
\end{table*}

\subsection{Main Results on WMT}\label{sec:WMT}

\begin{table}
\setlength\tabcolsep{4pt}
\centering
\resizebox{\columnwidth}{!}{
\begin{tabular}{lrrrrrr}
\hline
Models & En-De & En-Fr & En-Ru & En-Zh & En-Kk & En-Gu\\
\hline
baseline & 18\% & 23\% & 11\% & 29\% & 57\% & 68\% \\
(+ LD) & 19\% & 25\% & 11\% & 24\% & 42\% & 52\% \\
\hline
$\Delta$ & +1\% & +2\% & -0\% & -5\% & -15\% & -14\% \\
\hline
\end{tabular}}
\caption{\label{tab:copying_ratio}
The copying ratio for each language pair of XLM baselines and LD model. The average of the ratios of two directions for a language pair is reported. The translations used to compute the ratios are the same as translations for BLEU used in Table \ref{tab:high_wmt_perf} and Table \ref{tab:low_wmt_perf}.
}
\end{table}

As the proposed LD is helpful to alleviate the copying problem in Multi30K experiments when the weight $\lambda_{LD}$ is moderate, we further conduct experiments on WMT datasets, which are much larger than Multi30K. We use $\lambda_{LD} = 1$ as default.\\
\\
\noindent \textbf{High-resource language pairs.}
We report the results on Table \ref{tab:high_wmt_perf} and average copying ratios for each language pair in Table \ref{tab:copying_ratio}. Firstly, we observe that there is a slight decrease in BLEU scores for En-De and En-Ru pair. Different from Table \ref{tab:multi30k_perf} where we see that the vanilla models suffer from the copying problem, the vanilla models in Table \ref{tab:high_wmt_perf} perform fairly well on En-De and En-Ru. The copying ratios of each pair are also below 20\%. We therefore speculate that \textbf{the size} and \textbf{complexity} of the training data can influence the effectiveness of the language discriminator, as it can easily distinguish the decoder outputs in Multi30K because the size is small and each sentence has a similar and simple structure.
The copying problem does not severely impact the BLEU scores of these language pairs when training on WMT data, presumably because of the much larger dataset sizes.
When the two languages are more distant, however,
the
copying problem 
can
occur even if considerable training data is there: XLM baseline has a copying ratio of 29\% on En-Zh pair. XLM (+LD) can improve results by 0.44 and 0.41 in En $\veryshortarrow$ Zh and Zh $\veryshortarrow$ En directions, and decrease the copying ratio by 5\%, which indicates that the LD loss can improve the translation  where the copying problem is obvious.\\
\\
\begin{table}
\setlength\tabcolsep{4pt}
\centering
\footnotesize
\begin{tabular}{lrrrr}
\hline
Models & En$\veryshortarrow$Kk & Kk$\veryshortarrow$En & En$\veryshortarrow$Gu & Gu$\veryshortarrow$En \\
\hline
XLM baseline (512) &0.80&\textbf{2.00}&0.60&0.60 \\
XLM baseline (1024) &1.80&1.59&2.12&0.54 \\
XLM (+ LD) & \textbf{2.03}&1.70&\textbf{3.55}&\textbf{0.64}\\
\hline
\end{tabular}
\caption{\label{tab:low_wmt_perf}
BLEU scores of the XLM baseline and the same model enhanced with the LD loss on low-resource language pairs. The scores of baseline (512) are copied from \citep{kimetal2020unsupervised}. Same as the setting for high-resource languages, we reproduced XLM with 1024-dim embeddings to obtain the scores for baseline (1024). 
}
\end{table}\\
\noindent \textbf{Low-resource language pairs.} En-Kk and En-Gu represent two very distant pairs that include low-resource languages.
We report the BLEU scores in Table \ref{tab:low_wmt_perf} and average copying ratios in Table \ref{tab:copying_ratio}. From the results, we first see that the performance of all considered UNMT systems is rather poor. This is because they are all distant pairs and unsupervised training cannot learn enough cross-lingual information. 
We find the copying problem overwhelming, with 57\% and 68\% copying ratios on En-Kk and En-Gu pair respectively. By using the proposed LD loss, we see a consistent increase in BLEU scores and an evident decrease in average copying ratios (15\% decrease on En-Kk and 14\% on En-Gu pair respectively). This shows the incorporation of LD loss can significantly alleviate the copying problem. On the other hand, we attribute the weak translation quality to the already poor performance of the vanilla UNMT models, which cannot be largely improved simply by alleviating the copying problem. Decreasing copying ratios does not necessarily lead to a correct translation. Because of the unsupervised nature of the task, it can still be extremely hard for the model to learn enough cross-lingual information that is useful to perform good translation. Table \ref{tab:kk_translation} shows some examples, we notice that XLM (+ LD) generates sentences in the correct language, but the semantics of the output sentences is not that related to the original ones, indicating that lower copying ratios do not necessarily induce better translation quality.
\begin{table*}
\resizebox{\textwidth}{!}{ 
\begin{tabular}{|l|c|c|c|}
\hline
Model        & Source input                                     & Model output                    & Reference output                                  \\ \hline
XLM baseline & \multirow{2}{*}{\begin{otherlanguage*}{russian}Негізі , менің қарсылығым жоқ .\end{otherlanguage*}} & \begin{otherlanguage*}{russian}Негізі , менің қарсылығым жоқ . \end{otherlanguage*}& \multirow{2}{*}{Actually , I have no objection .} \\ \cline{1-1} \cline{3-3}
XLM (+LD)    &                                                  & \begin{otherlanguage*}{russian}" Негізі , I have no idea . \end{otherlanguage*}    &                                                   \\ \hline
XLM baseline & \multirow{2}{*}{\begin{otherlanguage*}{russian}Бұл сома алты еуроға тең .\end{otherlanguage*}} & \begin{otherlanguage*}{russian}The сома алты еуроға тең . \end{otherlanguage*}& \multirow{2}{*}{This amount equals to six euro .} \\ \cline{1-1} \cline{3-3}
XLM (+LD)    &                                                  & \begin{otherlanguage*}{russian} The price of six еуроға тең . \end{otherlanguage*}    &                                                   \\ \hline
XLM baseline & \multirow{2}{*}{\begin{otherlanguage*}{russian}\begin{tabular}[c]{@{}c@{}}Олардың көпшілігі ауыл шаруашылығы \\ саласында болып отыр .\end{tabular}\end{otherlanguage*}} & \begin{otherlanguage*}{russian}Their көпшілігі family life has changed . \end{otherlanguage*}& \multirow{2}{*}{Most of them are in agricultural area .} \\ \cline{1-1} \cline{3-3}
XLM (+LD)    &                                                  & \begin{otherlanguage*}{russian}\begin{tabular}[c]{@{}c@{}}Their family members have been \\ in the area for the past two years .\end{tabular} \end{otherlanguage*}    &                                                   \\ \hline
\end{tabular}
}
\caption{\label{tab:kk_translation}
Examples of translations from Kazakh to English by XLM baseline (1024) and XLM (+LD) in Table \ref{tab:low_wmt_perf}. The examples show XLM (+LD) suffers fewer the copying problem but it can generate incorrect tokens that do not match the semantics of the input sentence.
}
\end{table*}\\

Based on the high- and low-resource translation experiments, our insights are as follows: the UNMT models can (easily) learn a lot of cross-lingual information on similar and high-resource languages and thus the copying problem is less obvious. Under such a case, additionally using LD loss can divert the focus of the training. However, on distant pairs involving low-resource languages, models would struggle to learn enough cross-lingual information and therefore the copying problem is obvious. In such a case, although involving LD loss cannot provide additional cross-lingual knowledge, it can alleviate the copying problem thus improving the performance to a certain extent.  

\section{Discussion}

From the Multi30K and WMT experiments, we verify the ability of the LD loss to alleviate the copying problem by showing consistently lower copying ratios. However, the performance in terms of BLEU scores on these two datasets shows slightly different trends: we improve translation quality on Multi30K a lot by reducing the copying ratios; whereas we do not see a prominent improvement on WMT even if copying ratios are largely reduced. This discrepancy can be explained as follows. Two main issues are preventing the model from achieving good performance: (1) lacking cross-lingual alignment information that is useful for learning translation (2) no clear guidance on which language to translate into. The experiments on the small dataset Multi30K indicate that issue (1) is not the major obstacle when two similar languages are considered, e.g., En and Fr. In such a case, it is the issue (2) that prevents the model from performing the actual translation. This is why large improvements are achieved by simply adding the LD loss when training a model on Multi30k (note that the language discriminator does not provide any additional cross-lingual information but only acts as an implicit supervision). In the case of distant language pairs including low-resource languages, e.g., En-Gu and En-Kk in our WMT experiments, both issues (1) and (2) prohibit the model from learning to translate accurately. Although the copying problem is alleviated, as shown in Table \ref{tab:kk_translation}, this does not guarantee a correct or even good translation quality. We therefore expect future research could explore using a more powerful baseline model, e.g., including static cross-lingual embeddings to improve the cross-linguality \citep{chronopoulouetal2021improving}, which might further improve the performance for distant language pairs including low-resource languages.

\section{Conclusion}
In this paper, we find that the copying problem in UNMT is closely related to the lack of constraints on the intermediate translation in the BT process. To address this issue, we propose an LD loss to give additional supervision to the first-time-step output vectors generated by the decoder in the BT process. We find that the method can alleviate the copying problem by correcting the wrong behavior in BT. In addition, through extensive experiments on different language pairs (including low-resource languages and distant pairs), we discover that the method can consistently improve the performance of distant language pairs.

\section{Limitations and Risks}

Our training schedule introduces a language discriminator loss to impose constraints on the intermediate translation in the back-translation period. The experimental results suggest that our method can alleviate the copying problem when the involved languages are distant language pairs or lack training data. However, for language pairs that are not distant, and especially high-resource languages, our model does not show improvement over the baseline. Due to time and resource limitations, we do not further explore whether the optimal weight for the language discriminator loss can have a connection with the size of the dataset and the involved language pairs. For example, for WMT En-De or En-Fr pairs, the languages are not distant language pairs and therefore we might obtain better results if the weights are slightly smaller. We believe that future research could explore this direction: to adapt the weight to different language pairs and the size of the training data. 
In addition, we do not conduct hyperparameter search for other hyperparameters,
instead
directly 
using
suggested values. 

In this work, we propose a novel training schedule that tries to address the copying problem, which is common among distant language pairs in 
UNMT.
We experiment with high-resource languages English, German, French, Russian and Chinese, and low-resource languages including Gujarati and 	Kazakh. The 
training
data we use 
is
monolingual text extracted from online newspapers and released for the WMT series of shared tasks. As far as we know, all the monolingual corpora do not contain any metadata and therefore it would be unlikely that anyone can use the concerned data to attribute to specific individuals.

\section*{Acknowledgements}
We would like to thank the anonymous reviewers.
This work was funded by the European Research Council (grant \#740516) and by the German Research Foundation (DFG, grant FR $2829$/$4$-$1$).
\bibliography{anthology,custom}
\bibliographystyle{acl_natbib}

\appendix

\section{Appendix}
\label{sec:appendix}

\subsection{Scores of Other Metrics}

In addition to BLEU scores, we also compute other scores in other metrics, such as \textsc{chrF} \citep{popovic2015chrf} in Table \ref{tab:high_wmt_perf_chrf} and Table \ref{tab:low_wmt_perf_chrf}, COMET \citep{reietal2020comet} in Table \ref{tab:high_wmt_perf_comet} and Table \ref{tab:low_wmt_perf_comet}, and confidence interval of BLEU scores \citep{koehn2004statistical} in Table \ref{tab:high_wmt_perf_stats1}, Table \ref{tab:high_wmt_perf_stats2} and Table \ref{tab:low_wmt_perf_stats}. The translations used for computing the scores are the same as the translations used to compute the BLEU scores in Table \ref{tab:high_wmt_perf} and Table \ref{tab:low_wmt_perf}.

To quantify the copying problem, we use the copying ratio proposed by \citet{liuetal2021copying}, which is defined as follows:
\begin{equation}
\text{Ratio} = \frac{\sum_{i=1}^{I} \text{count(copying tokens)}}{\sum_{i=1}^{I}\text{count(tokens)}}
\end{equation}
where $I$ denotes the number of the total sentences in the test set, copying tokens are those tokens in the translation which are directly copied from the source language and the denominator is the total number of tokens in the generated translations. This metric will directly reflect the degree of the copying behavior of the translation model. The higher the copying ratio, the model tends to perform more copying instead translation. We report the average of the copying ratios of the two translation directions for each language pair in Table \ref{tab:copying_ratio}. We could see that the copying problem of the XLM baseline models is very obvious in low-resource language pairs, i.e., En-Kk and En-Gu. When the language discriminator loss is introduced, the copying ratios decrease by more than 10\%. We also notice that XLM (+LD) has a less obvious copying problem than the baseline in En-Zh pair, a distant language pair. For other language pairs, the copying problem is not that severe and therefore introducing the language discriminator loss does not much change the ratios.

\subsection{Model Details}\label{hyperparam}

In Section \ref{sec:multi30k}, we use the pretrained XLM models from HuggingFace\footnote{\url{https://github.com/huggingface}} \citep{wolf2020transformers} (\textit{xlm-mlm-enfr-1024}, \textit{xlm-mlm-ende-1024}) to initialize a shared encoder and randomly initialize a shared decoder. A single embedding layer (containing the words/subwords of both the source and target languages) from the pretrained encoder is used. The weight of the final fully connected layer is tied with the embedding layer. The parameters of the encoder are fixed except for this embedding
layer which is also used by the decoder. The embedding size is 1024 and the hidden size of the decoder is 512. The decoder has 8 heads and 3 layers. We follow the denoising autoencoding hyperparameter settings used by \citet{LampleCDR18umtmco} and the training schedule of \citet{liuetal2022flow}, i.e., firstly fine-tuning the models with only DAE loss and LD loss for the language discriminator for the first 2 epochs, then fine-tuning the models with all losses
(including the BT) for the rest of the epochs. We set the batch size to 32 and use Adam optimizer \citep{kingma2015adam} with an initial learning rate of 0.0001. We stop the training when the model does not improve the BLEU scores on the validation set for 5 epochs. We do not use beam search to generate translations for Multi30K.

In Section \ref{sec:WMT}, we pretrain all our own cross-lingual language models of each language pair based on XLM code base\footnote{\url{https://github.com/facebookresearch/XLM}}  \citep{Conneau2019xlm}. Then the encoder and decoder are both initialized with the same cross-lingual pretrained model. The recommended hyperparameters for the model architecture are used, i.e., 1024 for the embedding size, 4096 for the hidden size, 8 heads and 6 layers for the transformer blocks. We follow the recommended pretraining as well as UNMT fine-tuning hyperparameters from XLM. We only change the hyperparameter \textit{tokens\_per\_batch} to 250 to adapt to small- or moderate memory GPUs. We generate the translations by using beam search of size 5. These translations are used to compute the scores in all the WMT-related experiments.

For the language discriminator, we simply use a feed-forward neural network (FFNN). The language discriminator has two hidden layers and each layer has the same dimension as the embedding, i.e., 1024, for both Multi30K and WMT-related experiments. The output dimension is two which corresponds to the number of language domains we want to classify into, as we have two languages involved in the training for each model. 

\begin{table}
\setlength\tabcolsep{4pt}
\centering
\begin{tabular}{lrrrr}
\hline
Models & En$\veryshortarrow$Kk & Kk$\veryshortarrow$En & En$\veryshortarrow$Gu & Gu$\veryshortarrow$En \\
\hline
XLM baseline  & 8.85 & 7.61  & 7.95 & 4.76 \\
XLM (+ LD) & \textbf{11.78} & \textbf{10.09} & \textbf{11.71} & \textbf{7.12} \\
\hline
\end{tabular}
\caption{\label{tab:low_wmt_perf_chrf}
\textsc{chrF} scores \citep{popovic2015chrf} of the XLM UNMT baseline as well as the XLM model with the language discriminator on low-resource language pairs (the translations used are the same as used in Table \ref{tab:low_wmt_perf} for BLEU scores).
}
\end{table}

\begin{table}
\setlength\tabcolsep{4pt}
\centering
\begin{tabular}{lrrrr}
\hline
Models & En$\veryshortarrow$Kk & Kk$\veryshortarrow$En & En$\veryshortarrow$Gu & Gu$\veryshortarrow$En \\
\hline
XLM baseline  & -1.41 & -1.10  & -1.40 & -1.90 \\
XLM (+ LD) & \textbf{-1.14} & \textbf{-1.04} & \textbf{-0.91} & \textbf{-1.68}\\
\hline
\end{tabular}
\caption{\label{tab:low_wmt_perf_comet}
COMET scores \citep{reietal2020comet} of the XLM UNMT baseline as well as the XLM model with the language discriminator on low-resource language pairs (the translations used are the same as used in Table \ref{tab:low_wmt_perf} for BLEU scores). We use \textit{wmt20-comet-da} model to evaluate the translations.
}
\end{table}

\begin{table*}
\centering
\begin{tabular}{lrrrrrrrr}
\hline
Models & En$\veryshortarrow$De & De$\veryshortarrow$En & En$\veryshortarrow$Fr & Fr$\veryshortarrow$En & En$\veryshortarrow$Ru & Ru$\veryshortarrow$En & En$\veryshortarrow$Zh & Zh$\veryshortarrow$En\\
\hline
XLM baseline & \textbf{45.09} & \textbf{48.20} & \textbf{44.99} & 49.93 & \textbf{34.75} & \textbf{38.56} & 16.11 & 19.08\\
XLM (+ LD) & 44.42 & \textbf{48.20} & 42.94 & \textbf{50.50} & 34.39 & 36.56 & \textbf{16.74} & \textbf{20.45} \\
\hline
\end{tabular}
\caption{\label{tab:high_wmt_perf_chrf}
\textsc{chrF} scores \citep{popovic2015chrf} of the XLM UNMT baseline as well as the XLM model with the language discriminator on high-resource language pairs (the translations used are the same as used in Table \ref{tab:high_wmt_perf} for BLEU scores). 
}
\end{table*}

\begin{table*}
\centering
\begin{tabular}{lrrrrrrrr}
\hline
Models & En$\veryshortarrow$De & De$\veryshortarrow$En & En$\veryshortarrow$Fr & Fr$\veryshortarrow$En & En$\veryshortarrow$Ru & Ru$\veryshortarrow$En & En$\veryshortarrow$Zh & Zh$\veryshortarrow$En\\
\hline
XLM baseline & \textbf{-0.19} & \textbf{-0.22} & \textbf{-0.04} & 0.19 & \textbf{-0.34} & \textbf{-0.22} & -0.43 & \textbf{-0.78}\\
XLM (+ LD) & -0.22 & -0.23 & \textbf{-0.04} & \textbf{0.21} & -0.37 & -0.33 & \textbf{-0.36} & -0.81 \\
\hline
\end{tabular}
\caption{\label{tab:high_wmt_perf_comet}
COMET scores \citep{reietal2020comet} of the XLM UNMT baseline as well as the XLM model with the language discriminator on high-resource language pairs (the translations used are the same as used in Table \ref{tab:high_wmt_perf} for BLEU scores). We use \textit{wmt20-comet-da} model to evaluate the translations.
}
\end{table*}

\begin{table*}
\centering
\begin{tabular}{lrrrr}
\hline
Models & En$\veryshortarrow$De & De$\veryshortarrow$En & En$\veryshortarrow$Fr & Fr$\veryshortarrow$En\\
\hline
XLM baseline & 20.53$\pm$0.59 & 25.96$\pm$0.66 & \textbf{22.85$\pm$0.72} & \textbf{25.89$\pm$0.57}\\
XLM (+ LD) & 20.42$\pm$0.61 & 25.84$\pm$0.63 & \textbf{21.18$\pm$0.76} & \textbf{26.92$\pm$0.59} \\
\hline
\end{tabular}
\caption{\label{tab:high_wmt_perf_stats1}
95\% confidence interval for the BLEU scores of the XLM UNMT baseline as well as the XLM model with the language discriminator on En-De and En-Fr pair (the translations used are the same as used in Table \ref{tab:high_wmt_perf} for BLEU scores). Differences between bold results are statistically significant under $p=0.05$. For the statistical test, we use paired bootstrap resampling \citep{koehn2004statistical}.
}
\end{table*}

\begin{table*}
\centering
\begin{tabular}{lrrrr}
\hline
Models & En$\veryshortarrow$Ru & Ru$\veryshortarrow$En & En$\veryshortarrow$Zh & Zh$\veryshortarrow$En\\
\hline
XLM baseline  & \textbf{14.08$\pm$0.48} & \textbf{16.93$\pm$0.51} & \textbf{6.34$\pm$0.34} & \textbf{4.28$\pm$0.28} \\
XLM (+ LD) & \textbf{13.48$\pm$0.45} & \textbf{16.11$\pm$0.51} & \textbf{6.80$\pm$0.37} & \textbf{4.69$\pm$0.31}\\
\hline
\end{tabular}
\caption{\label{tab:high_wmt_perf_stats2}
95\% confidence interval for the BLEU scores of the XLM UNMT baseline as well as the XLM model with the language discriminator on En-Ru and En-Zh pair (the translations used are the same as used in Table \ref{tab:high_wmt_perf} for BLEU scores). Differences between bold results are statistically significant under $p=0.05$. For the statistical test, we use paired bootstrap resampling \citep{koehn2004statistical}.
}
\end{table*}

\begin{table*}
\centering
\begin{tabular}{lrrrr}
\hline
Models & En$\veryshortarrow$Kk & Kk$\veryshortarrow$En & En$\veryshortarrow$Gu & Gu$\veryshortarrow$En \\ 
\hline
XLM baseline  & 1.80$\pm$0.37 & 1.58$\pm$0.48 & \textbf{2.13$\pm$0.31} & 0.54$\pm$0.17 \\
XLM (+ LD) & 2.04$\pm$0.45 & 1.69$\pm$0.49 & \textbf{3.56$\pm$0.41} & 0.64$\pm$0.20\\
\hline
\end{tabular}
\caption{\label{tab:low_wmt_perf_stats}
95\% confidence interval for the BLEU scores of the XLM UNMT baseline as well as the XLM model with the language discriminator on En-Kk and En-Gu pair (the translations used are the same as used in Table \ref{tab:high_wmt_perf} for BLEU scores). Differences between bold results are statistically significant under $p=0.05$. For the statistical test, we use paired bootstrap resampling \citep{koehn2004statistical}.
}
\end{table*}

\end{document}